\pdfoutput=1

\documentclass[11pt]{article}

\usepackage[]{emnlp2021}

\usepackage{times}
\usepackage{latexsym}

\usepackage[T1]{fontenc}
\usepackage[utf8]{inputenc}

\usepackage{microtype}

\usepackage{amsmath,amsfonts,bm}

\def\eqref#1{equation~\ref{#1}}

\def\1{\bm{1}}

\DeclareMathAlphabet{\mathsfit}{\encodingdefault}{\sfdefault}{m}{sl}
\SetMathAlphabet{\mathsfit}{bold}{\encodingdefault}{\sfdefault}{bx}{n}

\usepackage{color}
\usepackage{amsmath}
\usepackage{amssymb}
\usepackage{multirow, varwidth}
\usepackage{dblfloatfix}
\usepackage{verbatim}
\makeatletter
\newif\if@restonecol
\makeatother

\usepackage[ruled,vlined]{algorithm2e}
\usepackage{xr}
\usepackage[amsmath,thmmarks]{ntheorem}

\DeclareRobustCommand\onedot{\futurelet\@let@token\@onedot}
\def\onedot{. }
\def\eg{\emph{e.g}\onedot} 
\def\ie{\emph{i.e}\onedot}

\usepackage{graphicx}
\usepackage{algorithmic}
\usepackage{mathtools}
\usepackage{caption}
\usepackage{booktabs}
\usepackage{multirow}
\usepackage{multirow}
\usepackage{url}
\usepackage{algorithm2e}
\usepackage{makecell}
\usepackage{lipsum}
\usepackage{soul}
\usepackage{tabularx}
\usepackage{adjustbox}
\usepackage{hyperref}

\newcolumntype{b}{X}
\newcolumntype{s}{>{\hsize=.5\hsize}X}

\usepackage{footnote}
\makesavenoteenv{tabular}
\makesavenoteenv{table}
\makesavenoteenv{minipage}
\makesavenoteenv{figure}

\definecolor{anger}{HTML}{FFC7BF}
\definecolor{joy}{HTML}{ffec99} %
\definecolor{excited}{HTML}{ffd8a8} %
\definecolor{faithful}{HTML}{ffd8a8} %
\definecolor{disgust}{HTML}{CFFFCC}
\definecolor{embarrassed}{HTML}{c3fae8} %
\definecolor{sadness}{HTML}{C0DEFF}
\definecolor{fear}{HTML}{d0bfff} %
\definecolor{grateful}{HTML}{bac8ff} %
\definecolor{surprised}{HTML}{faa2c1} %
\definecolor{selected}{HTML}{d3f9d8} %
\definecolor{guilty}{HTML}{adb5bd} %
\definecolor{proud}{HTML}{91a7ff} %
\definecolor{confident}{HTML}{e599f7} %

\newcommand{\hlc}[2][yellow]{{%
    \colorlet{foo}{#1}%
    \sethlcolor{foo}\hl{#2}}%
}

\newcommand{\joy}[1]{\hlc[joy]{#1}}
\newcommand{\angry}[1]{\hlc[anger]{#1}}

\newcommand{\fear}[1]{\hlc[fear]{#1}}
\newcommand{\sad}[1]{\hlc[sadness]{#1}}
\newcommand{\surprised}[1]{\hlc[surprised]{#1}}
\newcommand{\embarrassed}[1]{\hlc[embarrassed]{#1}}
\newcommand{\terrified}[1]{\hlc[fear]{#1}}
\newcommand{\guilty}[1]{\hlc[guilty]{#1}}
\newcommand{\faithful}[1]{\hlc[faithful]{#1}}
\newcommand{\anticipating}[1]{\hlc[joy]{#1}}
\newcommand{\proud}[1]{\hlc[proud]{#1}}
\newcommand{\confident}[1]{\hlc[confident]{#1}}
\newcommand{\sel}[1]{\hlc[selected]{#1}}

\definecolor{myred}{RGB}{255, 86, 93}
\definecolor{myblue}{RGB}{86, 125, 255}
\definecolor{urlcolor}{HTML}{E31C79} %

\title{Perspective-taking and Pragmatics for Generating \\ Empathetic Responses Focused on Emotion Causes}

\author{
    Hyunwoo Kim \qquad Byeongchang Kim \qquad Gunhee Kim \\
    Department of Computer Science and Engineering\\
    Seoul National University, Seoul, Korea \\
    {\tt \small{ \{hyunw.kim, byeongchang.kim\}@vl.snu.ac.kr gunhee@snu.ac.kr }} \\
    \href{https://vl.snu.ac.kr/projects/focused-empathy}{\textcolor{urlcolor}{\small{\texttt{https://vl.snu.ac.kr/projects/focused-empathy}}}}
}

\date{}

\begin{document}
\maketitle
\begin{abstract}
    Empathy is a complex cognitive ability based on the reasoning of others' affective states.
    In order to better understand others and express stronger empathy in dialogues, we argue that two issues must be tackled at the same time:
    (i) identifying which word is the cause for the other's emotion from his or her utterance and (ii) reflecting those specific words in the response generation.
    However, previous approaches for recognizing emotion cause words in text require sub-utterance level annotations, which can be demanding.
    Taking inspiration from social cognition, we leverage a generative estimator to infer emotion cause words from utterances with no word-level label.
    Also, we introduce a novel method based on pragmatics to make dialogue models focus on targeted words in the input during generation.
    Our method is applicable to any dialogue models with no additional training on the fly.
    We show our approach improves multiple best performing dialogue agents on generating more focused empathetic responses in terms of both automatic and human evaluation.
\end{abstract}

\section{Introduction}
\label{sec:intro}

Empathy is one of the hallmarks of social cognition.
It is an intricate cognitive ability that requires high-level reasoning on other's affective states.
The intensity of expressed empathy varies depending on the depth of reasoning.
According to \citet{Sharma:2020:EMNLP}, weak empathy is accompanied by generic expressions such as ``\textit{Are you OK?}'' or ``\textit{It's just terrible, isn't it?}'',
while stronger empathy reflects the other's specific situation:
``\textit{How is your \sad{headache}, any better?}'' or ``\textit{You must be worried about the \fear{job interview}}''.
In order to respond with stronger empathy, two issues must be tackled: %
reasoning (i) where to focus on the interlocutor's utterance (for the reason behind the emotion) and
(ii) how to generate utterances that focus on such words.

Firstly, which words should we focus on when empathizing with others?
As empathy relates to other's emotional states, the reasons behind emotions (\textit{emotion cause}) should be identified.
Imagine you are told ``\textit{I got a \joy{gift} from a \joy{friend} last vacation!}'' with a joyful face.
The likely words that can be the causes of his/her happiness are ``\textit{gift}'' and ``\textit{friend}''.
On the other hand, ``\textit{vacation}'' has less to do with the emotion.
If you respond ``\textit{How was your vacation?}'', the interlocutor may think you are not interested; %
rather, it is better to say ``\textit{Wow, what was the \joy{gift}?}'' or ``\textit{Your \joy{friend} must really like you.}'' by focusing on the emotion cause words.

We humans do not rely on word-level supervision for such affective reasoning.
Instead, we put ourselves in the other's shoes and simulate what it would be like.
\textit{Perspective-taking} is this act of considering an alternative point of view for a given situation.
According to cognitive science, \textit{perspective-taking} and \textit{simulation} are key components in empathetic reasoning \citep{Davis:1983:JPSP, Batson:1991:JPSP, Ruby:2004:JCogNeuro}.
Taking inspiration from these concepts, we propose to train a generative emotion estimator for simulating the other's situation and identifying emotion cause words.

Secondly, after reasoning which words to focus on, the problem of how to generate focused responses still remains.
Safe responses that can be adopted to any situations might hurt other's feelings.
Generated utterances need to convey the impression that concerns the specific situation of the interlocutor.
Such communicative reasoning is studied in the field of computational pragmatics.
The Rational Speech Acts (RSA) framework \citep{Frank:2012:Science} formulates communication between speaker and listener as probabilistic reasoning.
It has been applied to many tasks to increase the informativeness of generated text grounded on inputs \citep{Andreas:2016:EMNLP, Fried:2018:NeurIPS, Cohn:2019:NAACL, Shen:2019:NAACL}.
That is, RSA allows the input to be more reflected in the generated output.

However, controlling the RSA framework to reflect specific parts of the input remains understudied.
We introduce a novel method for the RSA framework to make models focus on targeted words in the interlocutor's utterance during generation.

In summary, we recognize emotion cause words in dialogue utterances with no word-level labels and generate stronger empathetic responses focused on them without additional training.
Our major contributions are as follows:

(1) We identify emotion cause words in dialogue utterances by leveraging a generative estimator.
Our approach requires no additional emotion cause labels other than the emotion label on the whole sentence, and outperforms other baselines.
(2) We introduce a new method of controlling the Rational Speech Acts framework \citep{Frank:2012:Science}
to make dialogue models better focus on targeted words in the input context to generate more specific empathetic responses.

(3) For evaluation, we annotate emotion cause words in emotional situations from the validation and test set of EmpatheticDialogues dataset \citep{Rashkin:2019:ACL}.
We publicly release our \textsc{EmoCause} evaluation set for future research.

(4) Our approach improves model-based empathy scores \citep{Sharma:2020:EMNLP} of three recent dialogue agents,
MIME~\citep{Majumder:2020:EMNLP}, DodecaTransformer~\citep{Shuster:2020:ACL}, and Blender~\citep{Roller:2021:EACL} on EmpatheticDialogues.
User studies also show that our approach improves human-rated empathy scores and is more preferred in A/B tests.

\section{Related Work}
\label{sec:related_work}

\textbf{Empathetic dialogue modeling.}
Incorporating user sentiment is one of early attempts for empathetic conversation generation \citep{Siddique:2017:ACL, Shi:2018:ACL}.
\citet{Rashkin:2019:ACL} collect a large-scale English empathetic dialogue dataset named EmpatheticDialogues. %
The dataset is now adopted in other dialogue corpus such as DodecaDialogue \citep{Shuster:2020:ACL} and BST \citep{Smith:2020:ACL}.
As a result, pretrained large dialogue agents such as DodecaTransformer \citep{Shuster:2020:ACL} and Blender \citep{Roller:2021:EACL} now show empathizing capabilities.
Empathy-specialized dialogue models are another stream of research.
Diverse architectures have been adopted, including emotion recognition \citep{Lin:2020:AAAI}, mixture of experts \citep{Lin:2019:EMNLP}, emotion mimicry \citep{Majumder:2020:EMNLP} and persona \citep{Zhong:2020:EMNLP}.
\citet{Li:2020:COLING} use lexicon to extract emotion-related words from utterances and feed them to a GAN-based agent.

We aim to improve both pretrained large dialogue agents and empathy-specialized ones by making them focus on emotion cause words in context.

\textbf{Emotion Cause (Pair) Extraction.}
The emotion cause extraction (ECE) task predicts causes in text spans, given an emotion.
Cause spans have been collected from Chinese microblogs and news \citep{Gui:2014:NLPCC, Gui:2016:EMNLP},
English novels \citep{Gao:2017:NTCIR}, and English dialogues \citep{Poria:2020:arxiv}.
\citet{Xia:2019:ACL} propose a task of extracting pairs of both emotion and its cause spans.
Previous works tackle these tasks via supervised learning with question-answering \citep{Gui:2017:EMNLP},
joint-learning \citep{Chen:2018:EMNLP}, co-attention \citep{Li:2018:EMNLP}, and regularization \citep{Fan:2019:EMNLP}.

Compared to those tasks, we recognize emotion cause words with no word-level labels using a generative estimator.
Our method does not require word-level labels other than the emotion labels of the whole sentences.
We then generate more specific empathetic responses focused on them.

\textbf{Rational Speech Acts (RSA) framework.}
The RSA framework \citep{Frank:2012:Science} has been applied to many NLP tasks including
referencing \cite{Andreas:2016:EMNLP, Zarriess:2019:ACL}, captioning \cite{Vedantam:2017:CVPR, Cohn:2018:NAACL},
navigating \cite{Fried:2018:NeurIPS}, translation \cite{Cohn:2019:NAACL}, summarization \cite{Shen:2019:NAACL}, and dialogue \citep{Kim:2020:EMNLP}.
It can improve informativeness of generated utterances better grounded on inputs (\eg images, texts).

Compared to previous use of RSA, we propose an approach that can control the models to focus on targeted words from the given input.

\section{Identifying Emotion Cause Words \newline with Generative Emotion Estimation}
\label{sec:gee}

Our approach consists of two steps: (i)  recognizing emotion cause words from utterances with no word-level labels (\S \ref{sec:gee}), and
(ii) generating empathetic responses focused on those words (\S \ref{sec:generation}).
In this section, we first train a generative emotion estimator to identify emotion cause words.

\subsection{Why Generative Emotion Estimator?}
\label{subsec:why_gee}
We leverage a \textit{generative} model by taking inspiration from \textit{perspective-taking} (\ie \textit{simulating} oneself in other's shoes) to reason emotion causes; not requiring word-level labels.
Our idea is to estimate the emotion cause weight of each word in the utterance while satisfying the following three desiderata.

(1) Do not require word-level supervision for learning to identify emotion cause words in the utterances.
Humans do not need word-level labels to infer the probable causes associated with the other's emotion during conversation.

(2) Simulate the observed interlocutor's situation within the model.
\textit{Simulation theory} (ST) from cognitive science explains that this mental imitation helps understanding the internal mental states of others \citep{Gallese:2004:TiCS}.
Much evidence for ST is found from neuroscience including mirror neurons \citep{Rizzolatti:2004:RevNeuro}, action-perception coupling \citep{Decety:2003:Neural}, and empathetic perspective-taking \citep{Ruby:2004:JCogNeuro}.

(3) Reason other's internal emotional states in Bayesian fashion.
Studies from cognitive science argue that human reasoning of other's affective states and minds can be described via Bayesian inference \citep{Griffiths:2008:Cambridge, Ong:2015:Cognition, Saxe:2017:Curr, Ong:2019:topics}.

Interestingly, a generative emotion estimator (GEE), which models $P(C,E) = P(E)P(C|E)$ with text sequence (\eg context) ${C}$ and emotion ${E}$, satisfies all the above conditions.
First, the generative estimator computes the likelihood of $C$ by \textit{generating} $C$ given $E$, which can be viewed as a \textit{simulation} of $C$.
Second, it estimates $P(E|C)$ via Bayes' rule.
Finally, the association between the emotion estimate and each word comes for free by using the likelihood of each words; without using any word-level supervision.
We use BART \citep{Lewis:2020:ACL} to implement a GEE. %

\subsection{Training to Model Emotional Situations}
\label{subsec:train_gee}

\textbf{Dataset}.
To train our GEE, we leverage the EmpatheticDialogues \citep{Rashkin:2019:ACL},
a multi-turn English dialogue dataset where the speaker talks about an emotional situation and the listener expresses empathy.
An example is shown in Table \ref{tab:ed_example}.
The emotion and the situation sentence are only visible to the speaker.
Situations are collected beforehand by asking annotators to recall related experiences for a given emotion label.
The dataset includes a rich suite of 32 emotion labels that are evenly distributed.

\textbf{Training}.
Given an emotion label $E$, GEE is trained to generate its corresponding emotional situation $C=\{w_1, ..., w_T\}$, where $w_i$ is a word.
As a result, our GEE learns the joint probability $P(C, E)$.
The trained GEE shows perplexity of 13.6 on the test situations of EmpatheticDialogues.

{\renewcommand{\arraystretch}{1.1}
    \begin{table}[t] \begin{center}
    \small
    \setlength{\tabcolsep}{1pt}
    \begin{tabularx}{\linewidth}{X}
        \toprule
        \textbf{Emotion}: Grateful \\
        \textbf{Situation}: \\
        I was grateful when my mother visited me for my birthday. \\
        \midrule
        \textbf{Speaker}: It was my birthday, my mom came to surprise me. \\
        \textbf{Listener}: Aw that's so nice, how did she surprise you? \\
        \textbf{Speaker}: She showed up to my house and brought me a cake. \\
        \textbf{Listener}: Cakes! yessss winning. :) \\
        \bottomrule
    \end{tabularx}
    \vspace{-7pt}
    \caption{A dialogue example in EmpatheticDialogues.} %
    \label{tab:ed_example}
    \vspace{-10pt}
\end{center}\end{table}}

{\renewcommand{\arraystretch}{1}
    \begin{table}[t] \begin{center}
    \small
    \setlength{\tabcolsep}{4pt}
    \begin{tabular}{l}
        \toprule
        \textbf{Emotion}: \hlc[joy]{Joyful} \\
        \textbf{GEE}:\\
        \textbullet~I got accepted into a masters program in neuroscience. \\
        \midrule
        \textbf{Emotion}: \hlc[anger]{Angry} \\
        \textbf{GEE}:\\
        \textbullet~I was so mad at my cousin. He stole my daughters stuff. \\
        \midrule
        \textbf{Emotion}: \hlc[grateful]{Grateful} \\
        \textbf{GEE}:\\
        \textbullet~The night my dad got me a new car was a magical time. \\
        \bottomrule
    \end{tabular}
    \caption{Example of sampled outputs from our generative emotion estimator (GEE) using Nucleus sampling.}
    \label{tab:gee_examples}
    \vspace{-15pt}
\end{center}\end{table}}

\subsection{Recognizing Emotions}
\label{subsec:gee_emotion}

Once trained, GEE can predict $P(E|C=c)$ for a word sequence $c$ (\eg utterance) using Bayes' rule: %
\begin{align}
    P(E|C=c) \propto P(C=c|E)P(E).
    \label{eq:emotion_recognition}
\end{align}
We compute the likelihood $P(C=c|E)$ by GEE's generative ability as described in \S \ref{subsec:why_gee}.
Since emotions in EmpatheticDialogues are almost evenly distributed, we set the prior $P(E)$ to a uniform distribution.
Finally, we find the emotion with the highest likelihood of the given sequence $c$. %

We comparatively report the emotion classification accuracy of GEE in Appendix.

\subsection{Weakly Supervised \newline Emotion Cause Word Recognition}
\label{subsec:gee_emotionalword}

We introduce how GEE can recognize emotion cause words solely based on emotion labels without word-level annotations.
For a given word sequence $c=\{w_1, w_2, ..., w_T\}$ (\eg utterance),
GEE can reason the association $P(W|E=\hat{e})$ of each word $w_t$ in the sequence $c$ to the recognized emotion $\hat{e}$ in Bayesian fashion:
\begin{align}
    P(W|E=\hat{e}) \propto P(E=\hat{e}|W)P(W).
    \label{eq:emotionalword_recognition}
\end{align}
The emotion likelihood is computed as 
\begin{align}
    P(\hat{e}|W=w_t) &= \mathbb{E}_{w_{<t}}[P(\hat{e}|w_t, w_{<t})] \\
                     & \approx \frac{P(w_t|\hat{e}, w_{<t})P(\hat{e})}{\sum_{e' \in \mathcal{E}} P(w_t|e', w_{<t})P(e')}, \notag
\end{align}
where $w_{<t}$ is the partial utterance up to time step $t-1$.
Since computing the expectation over all possible partial utterance $w_{<t}$ is intractable, we approximate it by a single sample.
We build set $\mathcal{E}$ to include $\hat{e}$ and emotions with the two lowest probability of $P(E|C=c)$  when recognizing emotion in Eq.(\ref{eq:emotion_recognition}). 
We assume the marginal $P(W)$ is uniform.
We choose the top-$k$ words reasoned by GEE as emotion cause words, and focus on them during empathetic response generation.
\section{Controlling the RSA framework for Focused Empathetic Responses}
\label{sec:generation}

\begin{figure}[t] \begin{center}
    \vspace{-5pt}
    \includegraphics[width=\linewidth]{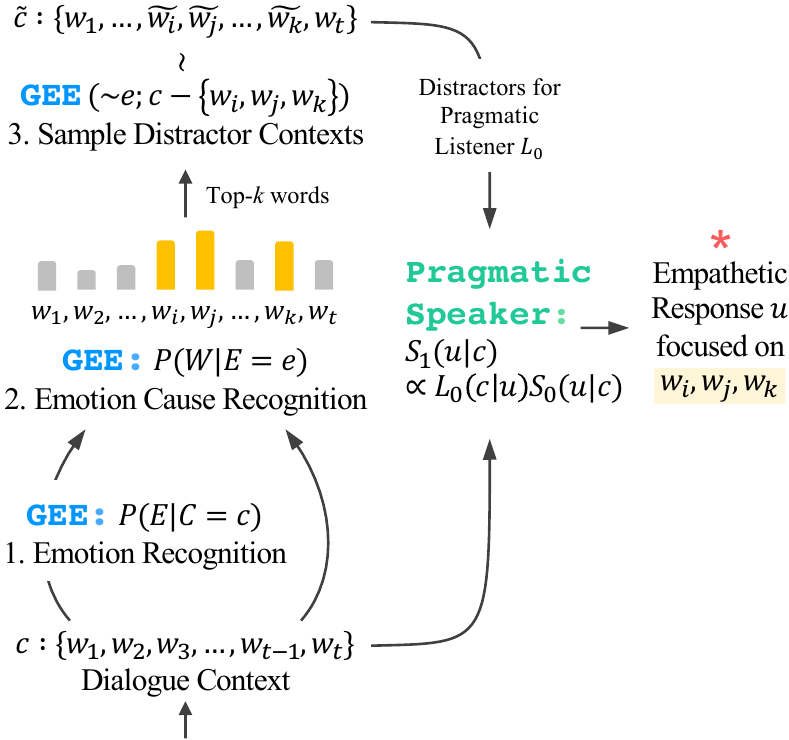}
    \caption{Overview of our method, consisting of emotion recognition (\S \ref{subsec:gee_emotion}),
        emotion cause word recognition (\S \ref{subsec:gee_emotionalword}),
        distractor context sampling (\S \ref{subsec:world}), and pragmatic generation (\S \ref{subsec:rsa}).
        GEE denotes our generative emotion estimator.}
    \label{fig:model}
    \vspace{-10pt}
\end{center} \end{figure}

We introduce how to control the Bayesian Rational Speech Acts (RSA) framework \citep{Frank:2012:Science} to focus on targeted words in the context during response generation.
We first preview the basics of RSA for dialogues (\S \ref{subsec:rsa}).
We then present how to control the RSA with word-level focus (\S \ref{subsec:world}), where our major contribution lies.
Figure \ref{fig:model} is the overview of our method.

\subsection{The Rational Speech Acts Framework}
\label{subsec:rsa}

Applying the RSA framework is computing the posterior of the dialogue agent's output distribution over words each time step. %
Hence, it is applicable to any existing pretrained dialogue agents on the fly, with no additional training.

The RSA framework formulates communication as a reference game between speaker and listener.
Based on recursive Bayesian formulation, the speaker (\ie dialogue model) reasons about the listener's belief of what the speaker is referring to.
We follow the approach of \citet{Kim:2020:EMNLP} for adopting RSA to dialogues.
Our goal here is to update a base speaker $S_0$ to a pragmatic speaker $S_1$ that focuses more on the emotion cause words in dialogue context $c$ (\ie dialogue history).

\textbf{Base Speaker $S_0$}.
Let $c$ and $u_t$ denote dialogue context and the output word of the model at time step $t$, respectively.
The base speaker $S_0$ is a dialogue agent that outputs $u_t$ for a dialogue context and partial utterance $u_{<t}$: $S_0(u_t|c, u_{<t})$.
As described, one can use any dialogue models for $S_0$.

\textbf{Pragmatic Listener $L_0$}.
The pragmatic listener is a posterior distribution over which dialogue context the speaker is referring to.
It is defined in terms of the base speaker $S_0$ and a prior distribution $p_t(C)$ over the context in Bayesian fashion:
\begin{multline} \label{eq:listener}
    L_0(c|u_{\leq t}, p_t) \\
    \propto \frac{S_0(u_t|c, u_{<t})^\beta \times p_t(c)}{\sum_{c' \in \mathcal{C}} S_0(u_t|c', u_{<t})^\beta \times p_t(c')}.
\end{multline}
The \textit{shared world} $\mathcal{C}$ is a finite set comprising the given dialogue context $c$ and other contexts (coined as \textit{distractors}) different from $c$.
Our contribution lies in how to build world $\mathcal{C}$ to endow the dialogue agent with controllability to better focus on targeted words, which we discuss in \S \ref{subsec:world}.
We update prior $p_{t+1}(C)$ with $L_0$ from time step $t$ as follows: $p_{t+1}(C) = L_0(C|u_{\leq t}, p_t)$.
$\beta$ is the rationality parameter which controls how much the base speaker's distribution is taken into account.
We note that $L_0$ is simply a distribution computed in Bayesian fashion, not another separate model.

\textbf{Pragmatic Speaker $S_1$}.
Integrating $L_0$ with $S_0$, we obtain the pragmatic speaker $S_1$:
\begin{multline}
    S_1(u_t|c, u_{<t}) \\
    \propto L_0(c|u_{\leq t}, p_t)^\alpha \times S_0(u_t|c, u_{<t}).
    \label{eq:s1}
\end{multline}
Since the pragmatic speaker $S_1$ is forced to consider how its utterance is perceived by the listener (via $L_0$),
it favors words that have high likelihood of the given context $c$ over other contexts in shared world $C$.
Similar to Eq. \ref{eq:listener}, $\alpha$ is the rationality parameter for $S_1$.
\subsection{Endowing Word-level Control for RSA \newline to Focus on Targeted Words in Context}
\label{subsec:world}

We aim to make dialogue models focus on targeted words from the input (\ie dialogue context) during generation via shared world $\mathcal{C}$.
The shared world $C$ consists of the given dialogue context $c$ and other distractor contexts.
It is used for computing the likelihood of the given context $c$ in Eq. \ref{eq:listener}.

Previous works of RSA in NLP manually (or randomly) select pieces of text (\eg sentences) entirely different from the given input \citep{Cohn:2019:NAACL, Shen:2019:NAACL, Kim:2020:EMNLP}.
In our context, it means distractors will be totally different contexts from $c$ in the dataset.
For example, when given a context ``\textit{I got a gift from my friend.}'', a distractor might be ``\textit{Today, I have an exam at school.}''.
Although such type of distractors helps improve the specificity of the model's generated outputs,
it is difficult to finely control which words the models should be specific about.

Our core idea is to build distractors by replacing the emotion cause words in $c$ with different words via sampling with GEE.
It can enhance the controllability of the RSA by making models focus on targeted words (\eg emotion cause words recognized by GEE) from the dialogue context.

For a dialogue context $c = \{w_1, ..., w_T\}$ where $w_i$ is a word, %
GEE outputs top-$k$ emotion cause words regarding the recognized emotion $\hat{e}_1$ from context $c$, denoted by $\mathcal{W}_{gee}$.
Next, we concatenate the least likely $n$ emotions from GEE with the context $c$ removing the top-$k$ emotion cause words:
$[\hat{e}_{-1}, ..., \hat{e}_{-n};c - \mathcal{W}_{gee}]$,
which is input to GEE.
We then sample different words ($\tilde{w}_i, \tilde{w}_j, \ldots, \tilde{w}_k$) from GEE's output in place of $\mathcal{W}_{gee}$ to construct a distractor $\tilde{c}$.
For example, given a context $c$ ``\textit{I was \sad{sick} from the \sad{flu}}'' and ``\textit{sick, flu}'' as the top-2 emotion cause words,
a sampled distractor $\tilde{c}$ can be ``\textit{I was \joy{laughing} from the \joy{relief}}''.
We use these altered contexts $\{\tilde{c}_1, ..., \tilde{c}_i\}$ as distractors for the shared world $\mathcal{C}$ in the pragmatic listener $L_0$ (Eq. \ref{eq:listener}).
We set $n$ and cardinality of world $\mathcal{C}$ to 3 (\ie $\mathcal{C} = \{c, \tilde{c}_1, \tilde{c}_2\}$).
We run experiments and find the best $k$ ($=5$) (see Appendix).

The only difference between the original context $c$ and the sampled distractor $\tilde{c}$ is those emotion cause words.
The pragmatic speaker $S_1$ (Eq. \ref{eq:s1}) prefers to generate words that have a
higher likelihood of the given context $c$ (including the original emotion cause words $\mathcal{W}_{gee}$) than the distractor context $\tilde{c}$. %
As a result, the pragmatic agent can generate utterances more focused on those original emotion cause words.

{\renewcommand{\arraystretch}{1.1}
    \begin{table}[t] \begin{center}
    \begin{adjustbox}{width=\columnwidth}
    \begin{tabular}{lcccc}
        \toprule
                                & \#Emotion      & Label     & \#Label/Utt    & \#Utt                 \\
        \midrule
        RECCON                  & 8              & Span      & 2.0             & 6.3K \\
        \textsc{EmoCause} (Ours)     & 32             & Word      & 2.3             & 4.6K \\
        \bottomrule
    \end{tabular}
    \end{adjustbox}
    \vspace{-5pt}
    \caption{
        Statistics of the \textsc{EmoCause} evaluation set compared to RECCON \citep{Poria:2020:arxiv}. Utt denotes utterance.
    }
    \vspace{-15pt}
    \label{tab:dataset_stats}
\end{center}\end{table}}

\begin{figure}[t] \begin{center}
    \includegraphics[width=\linewidth]{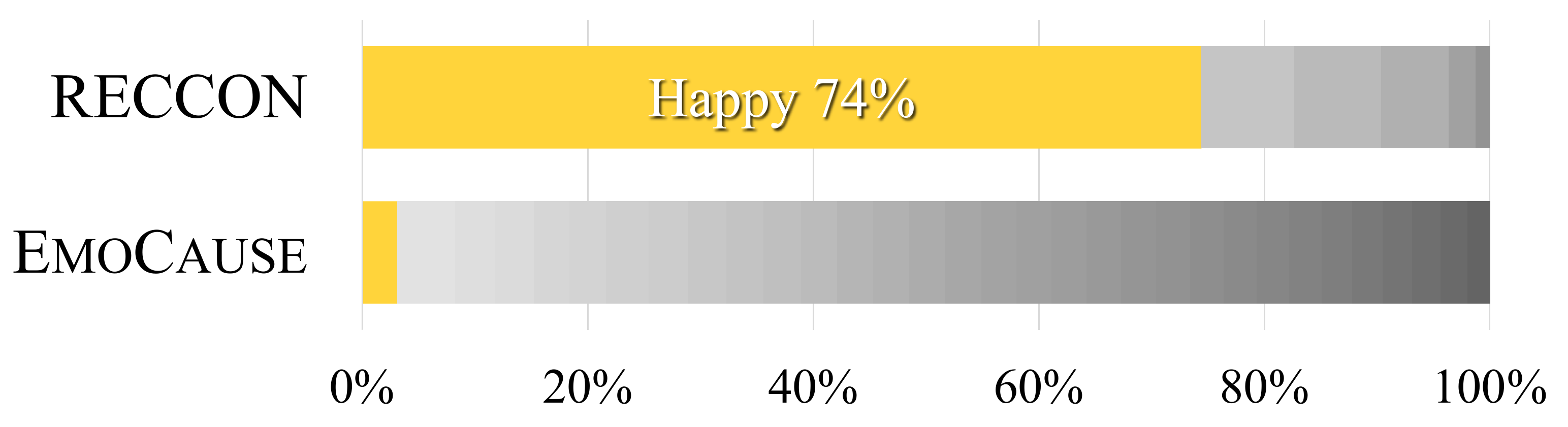}
    \vspace{-17pt}
    \caption{Emotion ratio of RECCON and our \textsc{EmoCause} evaluation set.}
    \label{fig:label_ratio}
    \vspace{-12pt}
\end{center} \end{figure}

{\renewcommand{\arraystretch}{1}%
    \begin{table}[t!] \begin{center}
    \small
    \setlength{\tabcolsep}{6pt}
    \begin{tabular}{cc}
        \toprule
        Emotion                               & Situation \\
        \midrule
        \addlinespace[0.2cm]
        \makecell{Surprised} & \makecell[l]{Man, I did not expect to \surprised{see} a \surprised{bear} on the \\ \surprised{road} today.} \\
        \addlinespace[0.1cm]
        \cmidrule(lr{0.3em}){1-2}
        \addlinespace[0.1cm]
        \makecell{Afraid}  & \makecell[l]{I have to take a \fear{business} \fear{trip} next week, \\
                                            I'm not looking forward to \fear{flying}.} \\
        \cmidrule(lr{0.3em}){1-2}
        \addlinespace[0.1cm]
        \makecell{Sad}  & \makecell[l]{I feel sad that I am \sad{spending} so much \sad{time} \\ this late on the \sad{internet}.} \\
        \cmidrule(lr{0.3em}){1-2}
        \addlinespace[0.1cm]
        \makecell{Joyful}  & \makecell[l]{I'm excited I get to \joy{go} to \joy{Disney} in October!} \\
        \addlinespace[0.1cm]
        \bottomrule
    \end{tabular}
    \vspace{-5pt}
    \caption{
        Examples of annotated emotion cause words.
    }
    \vspace{-10pt}
    \label{tab:dataset_example}
\end{center}\end{table}}

{\renewcommand{\arraystretch}{1}%
    \begin{table}[t!] \begin{center}
    \small
    \setlength{\tabcolsep}{4.5pt}
    \begin{tabular}{ll}
        \toprule
        {Embarrassed} & \makecell[c]{pant, fell, dropped, people, tripped, toilet} \\
        \cmidrule(lr{0.1em}){1-2}
        {Nostalgic} & \makecell[c]{old, childhood, memory, friend, back} \\
        \cmidrule(lr{0.1em}){1-2}
        {Trusting} & \makecell[c]{friend, gave, best, daughter, money, phone} \\
        \cmidrule(lr{0.1em}){1-2}
        {Anxious} & \makecell[c]{job, interview, exam, new, presentation} \\
        \cmidrule(lr{0.1em}){1-2}
        {Proud} & \makecell[c]{graduated, daughter, college, son, school} \\
        \cmidrule(lr{0.1em}){1-2}
        {Disappointed} & \makecell[c]{not, son, car, failed, get, job, hard, friend} \\
        \bottomrule
    \end{tabular}
    \vspace{-5pt}
    \caption{
        The most frequent cause words for each emotion. Other emotions can be found in Appendix.
    }
    \vspace{-15pt}
    \label{tab:frequent_words}
\end{center}\end{table}}

\section{\textsc{EmoCause}: \newline Emotion Cause Words Evaluation Set} %
\label{sec:dataset}

\subsection{Collecting Annotations}

To evaluate the performance of GEE, we annotate emotion cause words\footnote{
    As existing works annotate emotion cause spans for a given emotion label, we also coin our annotations as emotion cause words.
    However, in terms of ``\textit{causality}'', we note that the \textit{true} cause of the given emotion can be annotated only by the original annotator of the emotion label.
}
in the situations of validation and test set in EmpatheticDialogues \citep{Rashkin:2019:ACL} (\S \ref{subsec:train_gee}).
Using Amazon Mechanical Turk, we ask three workers to vote which words (\eg object, action, event, concept) in the situation sentence are the cause words to the given emotion.
Since explicit emotion words in the text (\eg happy, disappointed) are not cause words of emotion, we discourage workers from selecting them.

Annotators are required to have a minimum of 1000 HITs, 95\% HIT approval rate, and be located at one of [AU, CA, GB, NZ, US].
We pay the annotators \$0.15 per description.
To further ensure quality, only annotators who pass the qualification test are invited to annotate.
Nevertheless, speculations for emotion causes are subjective and can vary among annotators.
Therefore, we use \textit{only} unanimously selected words (\ie earning \textit{all} three votes) to ensure maximum objectivity.

\subsection{Analysis}

We analyze the characteristics of our emotion cause words in the \textsc{EmoCause} evaluation set.
In Table \ref{tab:dataset_stats} and Figure \ref{fig:label_ratio}, we compare the basic statistics of our annotation set and RECCON \citep{Poria:2020:arxiv}, %
which is an English dialogue dataset annotating emotion cause spans on the DailyDialog \citep{Li:2017:IJCNLP} and IEMOCAP \citep{Busso:2008:LREC} with a total of 8 emotions. %
Since our \textsc{EmoCause} is based on emotional situations from an empathetic dialogue dataset \citep{Rashkin:2019:ACL}, emotion causes play a more important role than in casual conversations from RECCON.
While 74\% of RECCON's labels belong to a single emotion \textit{happy}, \textsc{EmoCause} provides a balanced range of 32 emotions labels.
Therefore, our evaluation set presents a wider variety than RECCON.
Table \ref{tab:dataset_example} shows some examples of the annotated emotion cause words.

Table \ref{tab:frequent_words} reports the most frequent cause words for some emotions.
We find ``embarrassing'' events happen frequently in \textit{toilets} and in front of \textit{people}.
``Proud''  and ``disappointed'' are closely related to \textit{children}.
Interestingly, \textit{phones} are associated with ``trusting'', which may be due to smartphones containing sensitive personal information. %
More examples and results can be found in Appendix.

\section{Experiments}
\label{sec:experiments}
We first evaluate our generative emotion estimator (GEE) on weakly-supervised emotion cause word recognition (\S \ref{subsec:weakly-supervised}).
We then show our new controlling method for the RSA framework can improve best performing dialogue agents to generate more empathetic responses by better focusing on targeted emotion cause words (\S \ref{subsec:empathetic}). %

\subsection{Datasets and Experiment Setting}

\textbf{EmpatheticDialogues} (ED) \citep{Rashkin:2019:ACL}.
This dataset is an English empathetic dialogue dataset with 32 diverse emotion types (\S \ref{subsec:train_gee}).
The task is to generate empathetic responses (\ie responses from the listener's side in Table \ref{tab:ed_example}) when only given the dialogue context (\ie history) without emotion labels and situation descriptions.
It contains 24,850 conversations partitioned into training, validation, and test set by 80\%, 10\%, 10\%, respectively.
We additionally annotate cause words for the given emotion for all situations in the validation and test set of EmpatheticDialogues (\S \ref{sec:dataset}).

\textbf{EmoCause} (\S \ref{sec:dataset}).
We compare our GEE with four methods that can recognize emotion cause words with no word-level annotations:
random, RAKE \citep{Rose:2010:rake}, EmpDG \citep{Li:2020:COLING}, and BERT \citep{Devlin:2019:NAACL}.
For random, we randomly choose words as emotion causes.
RAKE is an automatic keyword extraction algorithm based on the word frequency and degree of co-occurrences.
EmpDG leverages a rule-based method for capturing emotion cause words using EmoLex \citep{Mohammad:2013:CI}, a large-scale lexicon of emotion-relevant words.
Finally, we train BERT for emotion classification with the emotion labels in ED.
For BERT, we select the words with the largest averaged weight of BERT's last attention heads for the classification token (\ie \texttt{[CLS]}).
More details can be found in Appendix.

\textbf{Dialogue models for base speakers}.
We experiment our approach on three recent dialogue agents: MIME \citep{Majumder:2020:EMNLP}, DodecaTransformer \citep{Shuster:2020:ACL}, and Blender \citep{Roller:2021:EACL}.
MIME is a dialogue model explicitly targeting empathetic conversation by leveraging emotion mimicry.
We select MIME, since it reportedly performs better than other recent empathy-specialized models \citep{Rashkin:2019:ACL, Lin:2019:EMNLP} on EmpatheticDialogues.
DodecaTransformer is a multi-task model trained on all DodecaDialogue tasks \citep{Shuster:2020:ACL} (\ie 12 dialogue tasks including ED, image and knowledge grounded ones) and finetuned on ED.
Blender is one of the state-of-the-art open domain dialogue agent \citep{Roller:2021:EACL} trained on BlendedSkillTalk dataset \citep{Smith:2020:ACL} which adopts contexts from ED.
We also finetune Blender on ED.
For all models, we use the default hyperparameters from the official implementations.
More details are in Appendix.

\textbf{Automatic evaluation metrics}.
For weakly-supervised emotion cause word recognition, we report the Top-$1,3,5$ recall scores. %

For EmpatheticDialogues, we report coverage and two scores for specific empathy expressions (Exploration, Interpretation) measured by pretrained empathy identification models \citep{Sharma:2020:EMNLP}.
The coverage score refers to the average number of emotion cause words included in the model's generated response.

The (i) Exploration and (ii) Interpretation are metrics for expressed empathy in text, introduced by \citet{Sharma:2020:EMNLP}.
They both require responses to focus on the interlocutor's utterances and to be specific.
(i) \underline{Explorations} are expressions of active interest in the interlocutor's situation, such as ``\textit{What happened?}'' or ``\textit{So, did you pass the chemistry exam?}''.
The latter is rated as a stronger empathetic response since it asks specifically about the interlocutor's situation.
(ii) \underline{Interpretations} are expressions of acknowledgments or understanding of the interlocutor's emotion or situation,
such as ``\textit{I know your feeling.}'' or ``\textit{I also had to speak in front of such audience, made me nervous.}''
Expressions of specific understanding are considered to be more empathetic.
RoBERTa models \citep{Liu:2019:arxiv} that are separately pretrained  for each metric rate each agent's response by returning values of 0, 1, or 2.
Higher scores indicate stronger empathy.

{\renewcommand{\arraystretch}{1.2}
    \begin{table}[t] \begin{center}
    \small
    \setlength{\tabcolsep}{11pt}
    \begin{tabular}{lccc}
        \toprule
        Model                                    & \makecell{Top-1\\Recall}     & \makecell{Top-3\\Recall}  & \makecell{Top-5\\Recall}    \\
        \midrule
        \makecell[l]{Human}                       & 41.3           & 81.1           & 95.0  \\
        \cmidrule(lr{1em}){1-4}
        \makecell[l]{Random}                      & 10.7           & 30.6           & 48.5  \\
        \makecell[l]{EmpDG}                       & 13.4           & 36.2           & 49.3  \\
        \makecell[l]{RAKE}                        & 12.7           & 35.8           & 55.0  \\
        \makecell[l]{BERT-Attention}              & 13.8           & 40.6           & 61.2  \\
        \makecell[l]{GEE (Ours)}                  & \textbf{17.3}  & \textbf{48.1}  & \textbf{68.4}  \\  %
        \bottomrule
    \end{tabular}
    \vspace{-5pt}
    \caption{
        Comparison of emotion cause word recognition performance between our generative emotion estimator (GEE),
        random, RAKE \citep{Rose:2010:rake}, EmpDG \citep{Li:2020:COLING}, and BERT
        on our \textsc{EmoCause} evaluation set (\S \ref{sec:dataset}). %
    }
    \vspace{-15pt}
    \label{tab:emotional_word}
\end{center}\end{table}}

\subsection{Weakly-Supervised \newline Emotion Cause Word Recognition}
\label{subsec:weakly-supervised}

Table \ref{tab:emotional_word} compares the recall of different methods on our \textsc{EmoCause} evaluation set (\S \ref{sec:dataset}).
Our GEE outperforms all other alternative methods.
RAKE performs better than EmpDG that uses a fixed lexicon of emotion-relevant words.
Compared to RAKE, methods leveraging dense word representations (\ie BERT, GEE) perform better.
Selecting words by BERT's attention weights does not attain better performance on capturing emotion cause words than GEE.
The gap between GEE and other methods widens when the number of returned words from models is more than one (\ie Top-$3,5$).

We also evaluate human performance to measure the difficulty of the task.
We randomly sample 100 examples from the test set and ask a human evaluator to select five best guesses for the emotion causes.
As the performance gap between GEE and human is significantly large, there is much room for further improvement in weakly-supervised emotion cause recognition.

\subsection{Empathetic Response Generation}
\label{subsec:empathetic}

{\renewcommand{\arraystretch}{1.1}%
    \begin{table}[t] \begin{center}
    \small
    \setlength{\tabcolsep}{4pt}
    \begin{tabular}{lccc}
        \toprule
        Model                         & \makecell{Coverage}  & \makecell{Exploration $\uparrow$}  & \makecell{Interpretation $\uparrow$} \\
        \midrule
        \multicolumn{4}{l}{MIME \cite{Majumder:2020:EMNLP}} \\
        \addlinespace[0.1cm]
        \hspace{1mm}$S_0$                 & 0.22                    & 0.12                                     & 0.05                                 \\
        \cmidrule(l{0.5em}){1-1}
        \hspace{1mm}Plain $S_1$           & 0.22                    & 0.23                                     & 0.10                                 \\
        \hspace{1mm}Focused $S_1$         & \textbf{0.24}           & \textbf{0.24}                            & \textbf{0.13}                         \\
        \midrule
        \multicolumn{4}{l}{DodecaTransformer \cite{Shuster:2020:ACL}}                                                                                                             \\
        \addlinespace[0.1cm]
        \hspace{1mm}$S_0$                & 0.34                      & 0.25                                     & 0.24                                 \\
        \hspace{1mm}$S_0$+Emotion        & 0.34                      & 0.21                                     & 0.20                                 \\
        \cmidrule(l{0.5em}){1-1}
        \hspace{1mm}Plain $S_1$          & 0.43                      & 0.30                                     & 0.23                                 \\
        \hspace{1mm}Focused $S_1$        & \textbf{0.49}             & \textbf{0.32}                            & \textbf{0.30}                                 \\
        \midrule
        \multicolumn{4}{l}{Blender \cite{Roller:2021:EACL}} \\
        \addlinespace[0.1cm]
        \hspace{1mm}$S_0$                & 0.35                      & 0.28                                     & 0.22                                 \\
        \hspace{1mm}$S_0$+Emotion        & 0.34                      & 0.31                                     & 0.20                                 \\
        \cmidrule(l{0.5em}){1-1}
        \hspace{1mm}Plain $S_1$          & 0.43                      & 0.37                                     & 0.21                                 \\
        \hspace{1mm}Focused $S_1$        & \textbf{0.54}             & \textbf{0.38}                            & \textbf{0.26}                                 \\
        \bottomrule
    \end{tabular}
    \vspace{-5pt}
    \caption{
        Comparison of our approach (Focused $S_1$) with other speakers on EmpatheticDialogues \cite{Rashkin:2019:ACL}.
        Exploration, and Interpretation scores are evaluated by pretrained RoBERTa models from \citet{Sharma:2020:EMNLP}.
    }
    \vspace{-10pt}
    \label{tab:ed_results}
\end{center}\end{table}}

\textbf{Results on Automatic Evaluation}.
Table \ref{tab:ed_results} reports the performance of different dialogue agents on EmpatheticDialogues \citep{Rashkin:2019:ACL} with automatic evaluation metrics.
Our \textit{Focused $S_1$} significantly outperforms the base model $S_0$ in terms of \underline{Interpretation} and \underline{Exploration} scores that measure more focused and specific empathetic expression.
We also test the plain pragmatic method (\textit{Plain $S_1$}) that use random distractors as in previous works \citep{Cohn:2018:NAACL, Kim:2020:EMNLP}.
The \textit{Focused $S_1$} consistently outperforms \textit{Plain $S_1$} on \underline{Interpretation} score with similar or better \underline{Exploration} scores.  %
The \textit{Focused $S_1$} models show higher coverage scores than other models, indicating they more reflect the context's emotion cause words in responses.
As MIME is only trained on EmpatheticDialogues, its \underline{Exploration} and \underline{Interpretations} scores are lower than models pretrained on other larger corpus.
As a result, we find our approach is effective in both large pretrained open domain dialogue models and empathy-specialized one.

{\renewcommand{\arraystretch}{1.1}%
    \begin{table}[t] \begin{center}
    \small
    \setlength{\tabcolsep}{6.9pt}
    \begin{tabular}{lccc}
        \toprule
        Model                       & Empathy $\uparrow$  & Relevance $\uparrow$  & Fluency $\uparrow$    \\
        \midrule
        \multicolumn{4}{l}{MIME \cite{Majumder:2020:EMNLP}} \\
        \addlinespace[0.1cm]
        \hspace{1mm}$S_0$             & 2.94            & 3.17                  & 2.75 \\
        \hspace{1mm}Focused $S_1$     & \textbf{3.09}   & \textbf{3.21}         & \textbf{2.83} \\
        \midrule
        \multicolumn{4}{l}{DodecaTransformer \citep{Shuster:2020:ACL}} \\
        \addlinespace[0.1cm]
        \hspace{1mm}$S_0$             & 2.53            & 3.47                  & 2.56                    \\
        \hspace{1mm}Focused $S_1$     & \textbf{2.71}   & \textbf{3.57}    & \textbf{2.75}                     \\
        \midrule
        \multicolumn{4}{l}{Blender \citep{Roller:2021:EACL}} \\
        \addlinespace[0.1cm]
        \hspace{1mm}$S_0$             & 2.91            & 3.12                  & 3.46                    \\
        \hspace{1mm}Focused $S_1$     & \textbf{3.00}   & \textbf{3.25}        & \textbf{3.57}                     \\
        \bottomrule
    \end{tabular}
    \vspace{-3pt}
    \caption{
        Comparison of our approach (Focused $S_1$) with base speakers $(S_0)$ on human rating. %
    }
    \vspace{-8pt}
    \label{tab:amt_human_rating}
\end{center}\end{table}}

{\renewcommand{\arraystretch}{1.1}%
    \begin{table}[t] \begin{center}
    \small
    \setlength{\tabcolsep}{10.6pt}
    \begin{tabular}{lccc}
        \toprule
        \hspace{-2mm}Model                                 & Win             & Lose          & Tie    \\
        \midrule
        \multicolumn{4}{l}{\hspace{-2mm}MIME \citep{Majumder:2020:EMNLP}} \\
        \addlinespace[0.1cm]
        \underline{Focused $S_1$} vs $S_0$     & \textbf{46.7}\%            & 20.0\%         & 33.3\%                     \\
        \midrule
        \multicolumn{4}{l}{\hspace{-2mm}DodecaTransformer \citep{Shuster:2020:ACL}} \\
        \addlinespace[0.1cm]
        \underline{Focused $S_1$} vs $S_0$     & \textbf{42.1}\%            & 28.8\%         & 29.1\%                     \\
        \midrule
        \multicolumn{4}{l}{\hspace{-2mm}Blender \citep{Roller:2021:EACL}} \\
        \addlinespace[0.1cm]
        \underline{Focused $S_1$} vs $S_0$    & \textbf{44.6}\%            & 37.4\%          & 18.0\%                     \\
        \bottomrule
    \end{tabular}
    \vspace{-4pt}
    \caption{
        Comparison of our approach (\textit{Focused $S_1$}) with base speakers $(S_0)$ on A/B test for empathetic response generation.
        The win and lose rates are based on \textit{Focused $S_1$}.
    }
    \vspace{-8pt}
    \label{tab:amt_ab_test}
\end{center}\end{table}}

{\renewcommand{\arraystretch}{1.1}%
    \begin{table}[t!] \begin{center}
    \small
    \setlength{\tabcolsep}{7.7pt}
    \begin{tabular}{lccc}
        \toprule
        \hspace{-2mm}Model                                 & Win             & Lose          & Tie    \\
        \midrule
        \multicolumn{4}{l}{\hspace{-2mm}Blender \citep{Roller:2021:EACL}} \\
        \addlinespace[0.1cm]
        \underline{Focused $S_1$} vs Plain $S_1$    & \textbf{39.6}\%            & 32.3\%          & 28.0\%                     \\
        \bottomrule
    \end{tabular}
    \vspace{-2pt}
    \caption{
        Comparison of our approach (\textit{Focused $S_1$}) with plain pragmatic approach (\textit{Plain $S_1$}) on A/B test for reflecting the given emotion cause words.
        The win and lose rates are based on \textit{Focused $S_1$}.
    }
    \vspace{-10pt}
    \label{tab:amt_ab_test_focused}
\end{center}\end{table}}

We also finetune DodecaTransformer and Blender with explicit emotion information ($S_0$+Emotion).
Following \citet{Rashkin:2019:ACL}, we concatenate the ground-truth emotion label to the dialogue context during training.
At inference, the top predicted emotion from GEE is used.
We find the \underline{Interpretation} or \underline{Exploration} scores of $S_0$+Emotion models drop.
Thus, simply adding emotion information is insufficient to make models focus more on the interlocutor's emotional event.

{\renewcommand{\arraystretch}{1.2}
    \begin{table}[t!] \begin{center}
    \small
    \begin{tabularx}{\linewidth}{X}
        \toprule
        \textbf{Emotion}: \angry{Angry} \\
        \textbf{Context}:
        I just moved to this neighborhood and some dumb criminals shot my neighbors and ran into the woods! \\
        \midrule
        \textbf{Emotion cause words from GEE}: \\
        dumb, neighbors, shot, moved, ran \\
        \midrule
        \textbf{$\mathbf{S_0}$}: That's terrible. Did you get a chance to talk to them?  \\
        \textbf{Plain $\mathbf{S_1}$}: \\ Oh my god, that is horrible! Where are they? \\
        \textbf{Focused $\mathbf{S_1}$}: \\ Oh my god, that is horrible! Did they get \underline{\textit{caught}}? \\
        \bottomrule
    \end{tabularx}
    \begin{tabularx}{\linewidth}{X}
        \toprule
        \textbf{Emotion}: \sel{Trusting} \\
        \textbf{Context}:
        I'm a big believer in listening to your gut. Sometimes I might wonder where my intuition is leading me, but it always seems to be spot on. \\
        \midrule
        \textbf{Emotion cause words from GEE}: \\
        believer, leading, intuition, wonder, spot \\
        \midrule
        \textbf{$\mathbf{S_0}$}: That's great. I'm glad you have a good intuition. \\
        \textbf{Plain $\mathbf{S_1}$}: \\ Good for you. I have a gut feeling about a lot of things. \\
        \textbf{Focused $\mathbf{S_1}$}: \\ That's great! I'm a big \underline{\textit{believer}} in \underline{\textit{intuition}} too. \\
        \bottomrule
    \end{tabularx}
    \begin{tabularx}{\linewidth}{X}
        \toprule
        \textbf{Emotion}: \proud{Proud} \\
        \textbf{Context}:
        We were living overseas at the times -- the most impressive part was crossing the Grand Canal over a pontoon of boats to finish the marathon ;-) \\
        \midrule
        \textbf{Emotion cause words from GEE}: \\
        impressive, Canal, Grand, part, finish \\
        \midrule
        \textbf{$\mathbf{S_0}$}: Wow, that sounds like a lot of fun! I'm glad you had a good time! \\
        \textbf{Plain $\mathbf{S_1}$}: \\ Wow, that sounds amazing! I bet you were so excited! \\
        \textbf{Focused $\mathbf{S_1}$}: \\ Wow, that sounds amazing! I've never been to \underline{\textit{Europe}}, but I've heard of it. \\
        \bottomrule
    \end{tabularx}
    \vspace{-5pt}
    \caption{
        Examples of recognized emotion cause words from our GEE and responses from $S_0$ and Focused $S_1$.
        We underline words where our Focused $S_1$ reflects the emotion cause words returned by GEE.\footnote{Since Grand Canal is a famous tourist attraction in Venice, Italy, the word `Europe' is closely related to it. We note that there is another famous Grand Canal in China. This might be a bias in BART, since it is trained on English datasets.}
    }
    \label{tab:generation_example}
\end{center}\end{table}}

\textbf{Results on Human Evaluation}.
We conduct user study and A/B test via Amazon Mechanical Turk.
We randomly sample 100 test examples, each rated by three unique human evaluators.
Following previous works \citep{Rashkin:2019:ACL, Lin:2019:EMNLP, Majumder:2020:EMNLP}, we rate empathy, relevance, and fluency of generated responses.
Given the dialogue context and model's generated response, evaluators are asked to rate each criterion in a 4-point Likert scale, where higher scores are better.
We also run human A/B test to directly compare the \textit{Focused} $S_1$ and base $S_0$.
We ask three unique human evaluators to vote which response is more empathetic.
They can select \textit{tie} if both responses are thought to be equal.

Table \ref{tab:amt_human_rating} and \ref{tab:amt_ab_test} summarizes the averaged human rating and A/B test results on MIME \citep{Majumder:2020:EMNLP}, DodecaTransformer \citep{Shuster:2020:ACL}, and Blender \citep{Roller:2021:EACL}.
Our \textit{Focused $S_1$} agents are rated more empathetic and relevant to the dialogue context than the base agent $S_0$, with better fluency.
Also, users prefer responses from our \textit{Focused $S_1$} agent over those from the base agent $S_0$.
The inter-rater agreement (Krippendorff's $\alpha$) for human rating and A/B test are 0.26 and 0.27, respectively; implying fair agreement.

In addition to the coverage score in Table \ref{tab:ed_results}, we run A/B test on Blender \citep{Roller:2021:EACL} to compare the \textit{Focused} $S_1$ and \textit{Plain} $S_1$ for reflecting the given emotion cause words in the responses.
We random sample 200 test examples and ask three unique human evaluators to vote which response is more focused on the given emotion cause words from the context.

Table \ref{tab:amt_ab_test_focused} is the result of A/B test for focused response generation on Blender \citep{Roller:2021:EACL}.
Users rate that responses from \textit{Focused $S_1$} more reflect the emotion cause words than those from the \textit{Plain $S_1$} approach.
Thus, both quantitative and qualitative results show that our \textit{Focused $S_1$} approach helps dialogue agents to effectively generate responses focused on given target words.

Examples of the recognized emotion cause words from GEE and generated responses are in Table \ref{tab:generation_example}.
Our \textit{Focused $S_1$} agent's responses reflect the context's emotion cause words returned from our GEE, implicitly or explicitly.

\section{Conclusion}

We studied how to use a generative estimator for identifying emotion cause words from utterances based solely on emotion labels without word-level labels (\ie weakly-supervised emotion cause word recognition).
To evaluate our approach, we introduce \textsc{EmoCause} evaluation set where we manually annotated emotion cause words on situations in EmpatheticDialogues \citep{Rashkin:2019:ACL}.
We release the evaluation set to the public for future research.
We also proposed a novel method for controlling the Rational Speech Acts (RSA) framework \citep{Frank:2012:Science} to make models generate empathetic responses focused on targeted words in the dialogue context.
Since the RSA framework requires no additional training, our approach is orthogonally applicable to any pretrained dialogue agents on the fly.
An interesting direction for future work will be reasoning how the interlocutor would react to the model's empathetic response.
Such reasoning is an essential part for expressing empathy.

\section*{Acknowledgments}

We thank the anonymous reviewers for their helpful comments.
This research was supported by Samsung Research Funding Center of Samsung Electronics under project number SRFCIT2101-01.
The compute resource and human study are supported by Brain Research Program by National Research Foundation of Korea (NRF) (2017M3C7A1047860).
Gunhee Kim is the corresponding author.

\bibliographystyle{acl_natbib}
\bibliography{emnlp2021_empathy}

\clearpage

\appendix

\section{Implementation Details}
\label{sec:implementation_detail}

\textbf{Weakly-supervised emotion cause word recognition}.
We use \textit{rake-nltk}\footnote{\url{https://github.com/csurfer/rake-nltk}} to implement RAKE \citep{Rose:2010:rake},
and the official code of EmpDG\footnote{\url{https://github.com/qtli/EmpDG}} from the authors \citep{Li:2020:COLING}.
We respectively finetune BERT-based-uncased \citep{Devlin:2019:NAACL} for BERT-Attention and BART-large \citep{Lewis:2020:ACL} for our generative emotion estimator (GEE).
We set a learning rate to 3e-5 for BERT-Attention and 1e-5 for GEE.
Other than the learning rate, we follow the default hyperparameters in ParlAI framework\footnote{\url{https://parl.ai}} \citep{Miller:2017:arxiv}.
We select the best performing checkpoint using the Top-1 recall for emotion cause word recognition on the validation set.
We run experiments 5 times with different random seeds and report averaged scores on Table \ref{tab:emotional_word}.

\textbf{Dialogue models}.
We use MIME \citep{Majumder:2020:EMNLP}, DodecaTransformer \citep{Shuster:2020:ACL}, and Blender 90M \citep{Roller:2021:EACL} as dialogue models for base speakers.
For MIME, we use the codes and pretrained weights of the authors' official implementation\footnote{\url{https://github.com/declare-lab/MIME}} as is.
For DodecaTransformer and Blender, we use the ParlAI framework with the default hyperparameters and finetune them on EmpatheticDialogues \citep{Rashkin:2019:ACL}.
We select the best performing checkpoint via perplexity on the validation set.

During inference, we use greedy decoding and set RSA parameter $\alpha$ and $\beta$ to 2.0 and 0.9 for MIME, 3.0 and 0.9 for DodecaTransformer, and 4.0 and 0.9 for Blender.
We select the best performing $\alpha$ and $\beta$ from the candidates of $[1.0, 2.0, 3.0, 4.0]$ and $[0.5, 0.6, 0.7, 0.8, 0.9, 1.0]$ with one trial for each.
Inference on the test set of EmpatheticDialogues takes 0.4 hours with Blender 90M base speaker.

\textbf{Evaluation metrics}.
To compute Exploration and Interpretation scores \citep{Sharma:2020:EMNLP}, we separately finetune RoBERTa-base for each score using the author's official code\footnote{\url{https://github.com/behavioral-data/Empathy-Mental-Health}}.

\textbf{Sensitivity to $k$ of top-$k$ emotion cause words}.
In all experiments, we use $k=5$, which is found by validation with $k=1,2,4,8$ using Blender \citep{Roller:2021:EACL} on EmpatheticDialogues \citep{Rashkin:2019:ACL}.
Table \ref{tab:topk_results} summarizes the results.

{\renewcommand{\arraystretch}{1}%
    \begin{table}[t!] \begin{center}
    \small
    \setlength{\tabcolsep}{11pt}
    \begin{tabular}{ccc}
        \toprule
        $k$                         & \makecell{Exploration $\uparrow$}  & \makecell{Interpretation $\uparrow$} \\
        \midrule
        \addlinespace[0.1cm]
        1                           & 0.32                                     & 0.27                                 \\
        2                           & 0.34                                     & 0.29                                 \\
        4                           & 0.35                                     & 0.30                         \\
        8                           & 0.36                                     & 0.29                         \\
        \bottomrule
    \end{tabular}
    \caption{
        Comparison of different $k$ values for top-$k$ emotion cause words on generating empathetic responses in EmpatheticDialogues \cite{Rashkin:2019:ACL}.
        Exploration and Interpretation scores are evaluated by pretrained RoBERTa models from \citet{Sharma:2020:EMNLP}.
    }
    \vspace{-7pt}
    \label{tab:topk_results}
\end{center}\end{table}}

Experiments for emotion cause word recognition and emotion classification are run on one NVIDIA Quadro RTX 6000 GPU.
Experiments for empathetic response generation are run on two GPUs.

\section{Emotion Classification}
\label{sec:emotion_classification}

We report the classification performance of emotion classifiers used in empathetic response generation.
Table \ref{tab:emo_classification} shows the Top-1, 5 emotion classification accuracy for each model.
For reference, BERT \citep{Devlin:2019:NAACL} shows 0.55 and 0.88 for Top-1 and 5 accuracy.

{\renewcommand{\arraystretch}{1}%
    \begin{table}[h!] \begin{center}
    \small
    \setlength{\tabcolsep}{11pt}
    \begin{tabular}{lcc}
        \toprule
        Model                               & \makecell{Top-1}  & \makecell{Top-5} \\
        \midrule
        \addlinespace[0.1cm]
        MoEL \citep{Lin:2019:EMNLP}           & 0.38               & 0.74                                 \\
        MIME \citep{Majumder:2020:EMNLP}      & 0.34               & 0.77                                 \\
        GEE (Ours)                            & 0.40               & 0.77                         \\
        \bottomrule
    \end{tabular}
    \caption{
        Comparison of emotion classification accuracy from different models trained on EmpatheticDialogues \cite{Rashkin:2019:ACL}.
    }
    \vspace{-10pt}
    \label{tab:emo_classification}
\end{center}\end{table}}

\section{Details of \textsc{EmoCause} Evaluation Set}
\label{sec:emotional_words_detail}

Table \ref{tab:annotation_examples_supp} shows some selected examples of emotion cause words with given emotion and situation.
Table \ref{tab:top10_frequent_words} shows Top-10 frequent cause words per emotion.
Interestingly, same words can be seen in both positive and negative emotions.
For example, we can find the word \textit{interview} on both ``Anxious'' and ``Confident''.
``Anticipating'' and ``Disappointed'' are closely related to \textit{vacation}.
This result shows that understanding the context is one of key prerequisites for emotion cause word recognition.

{\renewcommand{\arraystretch}{1.4}%
    \begin{table}[h!] \begin{center}
    \small
    \begin{tabularx}{\linewidth}{X}
        \toprule
        \textbf{Emotion}: Surprised \\
        We just got a \surprised{new} \surprised{puppy} . My older dog knew to let that one out first when I get home from work . \\
        \midrule
        \textbf{Emotion}: Faithful \\
        My \faithful{boyfriend} is going out with a bunch of people I do n't know tonight . But I trust him that he will be a \faithful{good} \faithful{boy} . \\
        \midrule
        \textbf{Emotion}: Anticipating \\
        I am really waiting on \anticipating{getting} my \anticipating{tax} \anticipating{returns} this year I could use new carpet \\
        \midrule
        \textbf{Emotion}: Trusting \\
        I trust my own \sel{intuitions} when it comes to my \sel{health} . \\
        \midrule
        \textbf{Emotion}: Embarrassed \\
        i was \embarrassed{super} \embarrassed{late} for my \embarrassed{meeting} on tuesday \\
        \midrule
        \textbf{Emotion}: Sad \\
        My girlfriend 's \sad{cat} is \sad{sick} with \sad{Cancer} . I do n't think she 's going to make it for much longer and I 'm really shaken up by it . \\
        \midrule
        \textbf{Emotion}: Proud \\
        I put in a lot of effort and \proud{energy} and I \proud{found} a \proud{new} \proud{job} . It 's an online teaching position and I feel so good about myself . \\
        \midrule
        \textbf{Emotion}: Terrified \\
        Driving down the highway during a heavy \terrified{thunderstorm} and a car \terrified{crash} happens in front of me where a car flips over . \\
        \midrule
        \textbf{Emotion}: Confident \\
        I \confident{studied} \confident{all} \confident{night} for my final exam \\
        \midrule
        \textbf{Emotion}: Guilty \\
        I made a really \guilty{inappropriate} \guilty{joke} about someone I work with to other coworkers and it got back to them . I feel really bad about it . \\
        \bottomrule
    \end{tabularx}
    \caption{
        Examples of our annotated emotion cause words.
        Words with background color are selected as emotion cause words by annotators.
    }
    \vspace{-10pt}
    \label{tab:annotation_examples_supp}
\end{center}\end{table}}

{\renewcommand{\arraystretch}{1}%
    \begin{table*}[t!] \begin{center}
    \small
    \begin{tabular}{lcl}
        \toprule
        \textbf{Emotion} & \textbf{\#Label/Utt} & \textbf{Top-10 frequent emotion cause words} \\
        \midrule
        Afraid        & 2.12 & alone, night, spider, house, noise, movie, dark, storm, hurricane, heard \\
        Angry         & 2.62 & car, dog, neighbor, friend, husband, brother, not, stole, hit, kid \\
        Annoyed       & 2.59 & dog, people, cat, work, loud, late, night, sister, neighbor, friend \\
        Anticipating  & 2.04 & new, waiting, vacation, coming, son, job, forward, next, friend, back \\
        Anxious       & 2.05 & interview, job, exam, presentation, big, dentist, going, test, girlfriend, back \\
        Apprehensive  & 2.11 & job, nervous, new, first, interview, driving, moving, car, day, night \\
        Ashamed       & 2.48 & stole, ate, friend, forgot, girlfriend, missed, drunk, bad, money, mistake \\
        Caring        & 2.49 & dog, sick, care, wife, friend, home, helped, puppy, girlfriend, baby \\
        Confident     & 1.95 & exam, studied, job, interview, win, test, well, prepared, good, answer \\
        Content       & 2.04 & life, good, happy, relaxing, watching, weekend, back, breakfast, family, live \\
        Devastated    & 2.42 & dog, passed, died, away, lost, friend, father, job, cancer, cat \\
        Disappointed  & 2.59 & not, son, car, failed, get, hard, job, n't, birthday, vacation \\
        Disgusted     & 2.47 & dog, poop, threw, friend, dead, food, roach, puked, eat, animal \\
        Embarrassed   & 2.73 & pant, fell, dropped, people, tripped, stuck, slipped, toilet, front, friend \\
        Excited       & 1.95 & vacation, new, friend, first, trip, car, puppy, see, won, coming \\
        Faithful      & 2.09 & loyal, girlfriend, husband, year, relationship, boyfriend, family, friend, married, good \\
        Furious       & 2.58 & car, dog, neighbor, hit, broke, without, son, room, accident, cheated \\
        Grateful      & 2.42 & friend, helped, life, job, family, good, help, husband, work, parent \\
        Guilty        & 2.64 & ate, stole, friend, forgot, money, candy, eating, cake, bar, girlfriend \\
        Hopeful       & 1.91 & job, promotion, future, new, better, get, interview, ticket, college, well \\
        Impressed     & 2.30 & friend, daughter, guy, car, new, well, man, brother, world, backflip \\
        Jealous       & 2.66 & friend, car, new, husband, girl, girlfriend, bought, got, boyfriend, won \\
        Joyful        & 2.18 & first, child, wife, friend, family, together, daughter, baby, birthday, trip \\
        Lonely        & 2.18 & friend, alone, moved, husband, family, myself, away, wife, went, left \\
        Nostalgic     & 2.59 & old, childhood, friend, memory, game, school, child, family, back, comic \\
        Prepared      & 2.00 & ready, packed, studied, exam, everything, supply, ingredient, studying, set, all \\
        Proud         & 2.40 & graduated, college, daughter, job, first, son, school, brother, won, new \\
        Sad           & 2.39 & dog, died, passed, away, cat, sick, friend, not, lost, put \\
        Sentimental   & 2.40 & old, picture, passed, photo, dog, childhood, school, away, toy, found \\
        Surprised     & 2.29 & friend, party, birthday, found, baby, car, gift, home, pregnant, won \\
        Terrified     & 2.28 & night, dog, tornado, car, bad, chased, someone, storm, fly, crash \\
        Trusting      & 2.17 & friend, best, daughter, drive, car, brother, sister, card, dog, phone \\
        \bottomrule
    \end{tabular}
    \caption{
        Number of emotion cause words per utterance and Top-10 frequent emotion cause words for each emotion.
    }
    \label{tab:top10_frequent_words}
\end{center}\end{table*}}

\end{document}